\pdfoutput=1

\documentclass[11pt]{article}

\usepackage[final]{acl}

\usepackage{times}
\usepackage{latexsym}

\usepackage[T1]{fontenc}
\usepackage[utf8]{inputenc}
\usepackage{algorithm}
\usepackage{algorithmic}
\usepackage{makecell}
\usepackage{amssymb}
\usepackage{hyperref}
\usepackage{color}
\definecolor{span_pink}{RGB}{255,122,179}
\definecolor{entity_blue}{RGB}{148,169,216}
\usepackage{newfloat}
\usepackage{listings}
\floatstyle{ruled}
\newfloat{listing}{tb}{lst}{}
\floatname{listing}{Listing}

\usepackage{microtype}
\usepackage{booktabs} 
\usepackage{url}
\usepackage{float}
\usepackage{graphicx}
\usepackage{tabularx}
\usepackage{multirow}
\usepackage{multicol}
\usepackage{balance}
\usepackage{amsfonts}
\usepackage{amsmath}
\usepackage{xcolor}
\usepackage{colortbl}
\usepackage{diagbox}
\usepackage{subfigure}
\makeatletter

\newtheorem{theorem}{Theorem}
\newtheorem{lemma}{Lemma}
\newtheorem{proof}{Proof}[section]

\newcommand{\thickhline}{%
    \noalign {\ifnum 0=`}\fi \hrule height 1pt
    \futurelet \reserved@a \@xhline
}
\newcommand*\circled[1]{\kern-2.5em%
  \put(0,4){\color{white}\circle*{18}}\put(0,4){\circle{10}}%
  \put(-3,0){\color{black}\bfseries#1}~~}
\usepackage{tikz} 

\usepackage{enumitem}

\newcommand{\printfnsymbol}[1]{%
  \textsuperscript{\@fnsymbol{#1}}%
}
\title{VLA-Mark: A cross modal watermark for large \\ vision-language alignment models}

\author{Shuliang Liu\textsuperscript{\rm 1,2}, Qi Zheng\textsuperscript{\rm 1,2},  Jesse Jiaxi Xu\textsuperscript{\rm 3}, Yibo Yan\textsuperscript{\rm 1,2}, \textbf{Junyan Zhang}\textsuperscript{\rm 1,2}, \textbf{He Geng}\textsuperscript{\rm 1,2} \\  \textbf{Aiwei Liu}\textsuperscript{\rm 1,2}, \textbf{Peijie Jiang}\textsuperscript{\rm 4}, \textbf{Jia Liu}\textsuperscript{\rm 4}, \textbf{Yik-Cheung Tam}\textsuperscript{\rm 5}, \textbf{Xuming Hu}\textsuperscript{\rm 1,2 }\thanks{~\ Corresponding author.}\\
        \textsuperscript{\rm 1} {The Hong Kong University of Science and Technology (Guangzhou)} \\
    { \textsuperscript{\rm 2} {The Hong Kong University of Science and Technology}} \\
    { \textsuperscript{\rm 3} {University of Toronto}}
    { \textsuperscript{\rm 4} {Ant Group, Alibaba}} 
    { \textsuperscript{\rm 5} {New York University Shanghai}}
    \\
     \texttt{\href{mailto:shulianglyo@gmail.com}{shulianglyo@gmail.com}},
     \texttt{\href{mailto:xuminghu@hkust-gz.edu.cn}{xuminghu@hkust-gz.edu.cn}}}

\begin{document}
\maketitle
\begin{abstract}

Vision-language models demand watermarking solutions that protect intellectual property without compromising multimodal coherence. Existing text watermarking methods disrupt visual-textual alignment through biased token selection and static strategies, leaving semantic-critical concepts vulnerable. We propose \textbf{VLA-Mark}, a vision-aligned framework that embeds detectable watermarks while preserving semantic fidelity through cross-modal coordination. Our approach integrates multiscale visual-textual alignment metrics, combining localized patch affinity, global semantic coherence, and contextual attention patterns, to guide watermark injection without model retraining. An entropy-sensitive mechanism dynamically balances watermark strength and semantic preservation, prioritizing visual grounding during low-uncertainty generation phases. Experiments show 7.4\% lower PPL and 26.6\% higher BLEU than conventional methods, with near-perfect detection (98.8\% AUC). The framework demonstrates 96.1\% attack resilience against attacks such as paraphrasing and synonym substitution, while maintaining text-visual consistency, establishing new standards for quality-preserving multimodal watermarking  \footnote{Code is available at \url{https://github.com/shiningwhite-cmd/VLA-mark}}. 
\end{abstract}

\section{Introduction}

The emergence of vision-language aligned multimodal large models (VLAMMs) has fundamentally transformed cross-modal content generation. Pioneering architectures like LLaVA~\cite{liuVisualInstructionTuning2023} and Flamingo~\cite{alayrac2022flamingo} establish joint embedding spaces through cross-modal attention mechanisms, enabling unprecedented visual-linguistic synergy. These models achieve state-of-the-art performance in vision-language tasks ranging from contextual image captioning to visual commonsense reasoning~\cite{li2025treehop}, with recent extensions like Mini-Gemini~\cite{li2024mini} demonstrating human-level multimodal comprehension. \cite{liu2024adaptive,yoo2024advancing,ling2025wakenllm} However, \textit{their rising capability to generate semantically coherent cross-modal content urgently demands robust solutions for intellectual property protection and content authenticity}.

Embedding imperceptible yet detectable watermarks into LLM-generated outputs has emerged as a pivotal solution, yet existing techniques predominantly focus on unimodal scenarios. The pioneering "green list" partitioning~\cite{kirchenbauer2023watermark} establishes fundamental watermarking frameworks through vocabulary bias induction, while subsequent improvements like unbiased probability of two partitioned lists~\cite{mao2024watermark} and distribution-preserving strategies~\cite{wuresilient} enhance quality-robustness trade-offs in text generation. However, \textit{these approaches fail to address the unique challenges of multimodal generation where visual semantics critically guide textual outputs}. 

Current watermarking methodologies exhibit three critical limitations when applied to vision-language aligned generation. First, traditional text watermarking approaches like "green list" partitioning~\cite{kirchenbauer2023watermark} disrupt vision-conditioned language generation by introducing vocabulary biases that contradict visual semantics - for instance, suppressing visually grounded entity mentions detected through region-based attention. Even advanced context-aware variants~\cite{ren2023robust} fail to account for cross-modal dependencies established through vision-language projection layers in models like BLIP-2~\cite{li2023blip}. Second, static watermark allocation strategies~\cite{liang2024watermarking,zhao2023provable} typically apply uniform injection intensities regardless of position-specific visual grounding strength, leading to disproportionate distortion of visually salient tokens. This limitation persists even in theoretically-grounded approaches~\cite{huang2023towards} that optimize statistical trade-offs but ignore entropy variations during cross-modal generation. Third, current methods lack explicit mechanisms to protect vision-critical semantics under text-space attacks. Random vocabulary partitioning and uniform logit manipulation render key visual concepts (e.g., objects, scene descriptors) vulnerable to adversarial paraphrasing or synonym substitution. As shown in Fig.~\ref{fig:overview} (5), conventional watermarks indiscriminately boost non-semantic tokens (green blocks) while leaving visually anchored phrases like "grassy trail" (light blue blocks) exposed to semantic erasure through token replacement attacks. This fundamentally undermines text-visual coherence and detection consistency.

We resolve these challenges through \textbf{VLA-Mark}, the \textbf{first vision-language aligned watermarking framework that achieves cross-modally coordinated, quality-preserving watermark with excellent detectability and robustness} via three innovations. First, extending beyond random vocabulary splitting, our \textit{Multiscale Semantic Saliency Metrics} leverage visual semantics to guide green list selection through localized patch affinity (LPA), global semantic coherence (GSC)~\cite{hu2022hiure}, and cross-modal contextual salience (CCS). This aligns token partitioning with image content while maintaining zero training overhead. Second, our \textit{Entropy-Regulated Partition} dynamically adjusts watermark intensity based on generation uncertainty and token criticality scores, prioritizing semantic preservation in low-entropy phases while enhancing watermark strength during high-entropy generation. Third, we introduce \textit{SCT based Distribution Adjustment} through vision-aligned token prioritization, where cross-modal embedding alignment and fused metrics establish hierarchical protection for \textbf{Semantic Critical Tokens} (SCTs) against textual perturbations.

Our contributions transcend prior art through three breakthroughs: 
\begin{itemize}
    \item We pioneer the first text watermarking method for vision-language models, achieving cross-modal semantic guidance through native alignment mechanisms of VLA architectures, yielding 7.4\% and 26.6\% average improvement (PPL↓ and BLEU↑) in textual quality with zero training overhead. 
    \item We develop an uncertainty-aware coordination mechanism that automatically adapts watermark intensity to logits entropy, breaking the preservation-detection trade-off by maintaining SOTA detection performance while enhancing generation quality. 
    \item Through dedicated SCT preservation, we establish hierarchical protection against Paraphrase, Synonym, Translate and more attacks, ensuring text-visual consistency under perturbations. 
\end{itemize}

\section{Methodology}

\begin{figure*}[t!]
    \centering
    \includegraphics[width=0.95\linewidth]{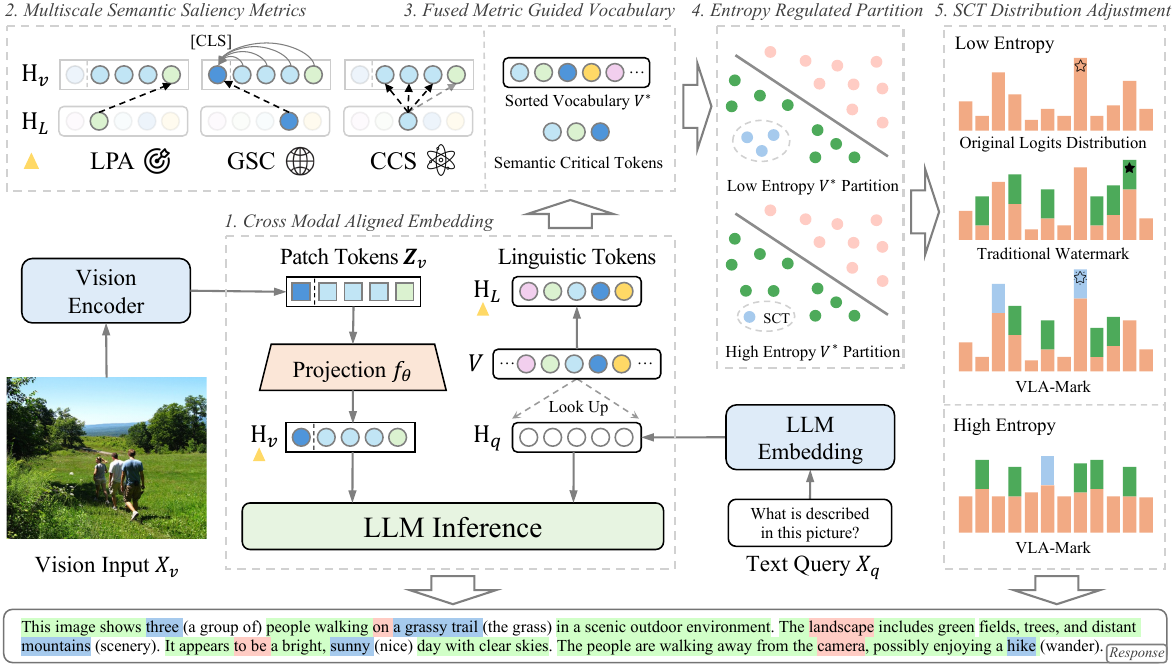}    
\caption{Proposed VLA-Mark framework. Vision embeddings $\mathbf{H}_v$ (aligned to LLM space) and linguistic tokens $\mathbf{H}_L$ extracted from LLM vocabulary $\mathcal{V}$ compute fused multiscale metrics (LPA/GSC/CCS) to rank $\mathcal{V}^*$ by visual saliency. Entropy-regulated SCT selection dynamically enhances semantic expressiveness when low entropy in logits distribution or watermark robustness when high entropy. Light blue \textcolor[HTML]{A6CAED}{\rule{0.2cm}{0.2cm}} denote SCT, which  in the response is followed by conventional watermarked tokens.}    
    \label{fig:overview}
\vspace{-2mm}
\end{figure*}

Our VLA-Mark framework introduces a vision-aligned watermarking method that identifies \textbf{Semantic Critical Tokens (SCTs)}, linguistic units strongly grounded in visual semantics guided by cross-modal embedding alignment (Sec~\ref{sec:crossmodal}) and fused multiscale metrics (Sec~\ref{sec:multiscale}). SCTs preserve text-visual coherence by anchoring key concepts (\textit{e.g.,} objects/scenes) while enabling entropy-regulated dynamic vocabulary partitioning (Sec~\ref{sec:dynamics}): low-entropy contexts prioritize SCT retention for semantic fidelity, whereas high-entropy phases emphasize watermark strength. The method further adjusts token distributions through watermarked logit manipulation (Sec~\ref{sec:adjustment}). This approach pioneers visual semantics as the foundation for watermark injection, contrasting traditional text-only statistical strategies, as is illustraed in Fig. \ref{fig:overview}.  For more theoretical analysis of each part, please refer to Appendix \ref{app:theoretical_analysis}.

\subsection{Cross-Modal Aligned Embedding} \label{sec:crossmodal}

As demonstrated in prior research, Vision-Language Alignment (VLA) models like LLaVA \cite{liuVisualInstructionTuning2023} employ a shared semantic mapping strategy where visual embeddings are projected into the text embedding space. 

Given a textual instruction $X_q$ and visual input $X_v$, such models utilize parallel encoding streams to process multimodal inputs. The vision encoder (\textit{e.g.,} SigLIP \cite{zhai2023sigmoid} or ViT-L/14 \cite{radford2021learning}) generates spatial-visual features through:
\begin{equation}
\begin{aligned}
\mathbf{Z}_v &= \text{VisEnc}(X_v) = [\mathbf{z}_{\text{cls}}; \mathbf{z}_1, ..., \mathbf{z}_P],
\end{aligned}
\end{equation}
where $\mathbf{Z}_v \in \mathbb{R}^{(P+1) \times d_v} $ and $P$ indicates the total number of image patch tokens augmented with a global [CLS] token. The subsequent alignment phase employs a trainable projection module $f_{\theta}(\cdot)$ , implemented as MLP \cite{liu2024improved} or generation adaptor \cite{chen2025januspro}, to bridge the dimensional gap between modalities:
\begin{equation}
\mathbf{H}_v = f_{\theta}(\mathbf{Z}_v) ,
\end{equation}

where $f_{\theta}$ denotes parametric transformation that enables cross-modal compatibility while retaining original information patterns, so we get $\mathbf{H}_v \in \mathbb{R}^{(P+1) \times d}$ . 
LLMs (\textit{e.g.,} Vicuna \cite{chiang2023vicuna}) first tokenize input text of length $S$ and then retrieve text embeddings $\mathbf{H}_q \in  \mathbb{R}^{S \times d}$ for LLM inference by querying the pretrained token embedding table, commonly referred to as the Vocabulary $\mathcal{V}$.
We construct an embedding matrix $\mathbf{H}_{L}$ by removing non-linguistic elements such as symbols and numbers from $\mathcal{V}$, where $L$ denotes the number of linguistic tokens in the vocabulary. Then we use $\mathbf{H}_{v}$ and $\mathbf{H}_{L}$ in the following modules to find the SCT to guided $\mathcal{V}$ partitioning for watermark.

\subsection{Multiscale Semantic Saliency Metrics} \label{sec:multiscale}
The $l$-th token embedding in $\mathbf{H}_{L}$ is denoted as $\mathbf{h}_{L}^{(l)}$.
We propose three complementary metrics to evaluate semantic criticality of linguistic tokens from orthogonal perspectives:

1. \textbf{Localized Patch Affinity (LPA)} quantifies region-specific importance by identifying the most relevant visual patch:
\begin{equation}
\psi_{\operatorname{LPA}}(l) = \max_{1 \leq p \leq P} \frac{\mathbf{h}_v^{(p)} \cdot \mathbf{h}_{L}^{(l)}}{\|\mathbf{h}_v^{(p)}\| \|\mathbf{h}_{L}^{(l)}\|}.
\end{equation}
\textbf{Role}: LPA captures \textit{fine-grained visual grounding} by measuring the maximum alignment between a text token and individual image regions. This is critical for detecting \textit{object-centric tokens} (e.g., "grassy trail", "mountain") that strongly correlate with localized visual patterns. However, it may underestimate tokens with \textit{diffuse visual associations} (e.g., "park", "crowded") that judged by the whole image.

2. \textbf{Global Semantic Coherence (GSC)} measures holistic alignment with the entire visual scene:
\begin{equation}
\psi_{\operatorname{GSC}}(l) = \frac{\mathbf{h}_v^{(\text{cls})} \cdot \mathbf{h}_{L}^{(l)}}{\|\mathbf{h}_v^{(\text{cls})}\| \|\mathbf{h}_{L}^{(l)}\|}.
\end{equation}
\textbf{Role}: GSC evaluates \textit{scene-level consistency} by comparing text tokens to the global visual representation ([CLS] token). It prioritizes tokens that summarize the scene (e.g., "sunny", "hike") or anchor high-level semantics. However, global pooling may dilute \textit{localized but critical details} come from certain patches (e.g., "broken" in a damaged object).

3. \textbf{Cross-Modal Contextual Salience (CCS)} aggregates multi-region visual relevance through attention weights:
\begin{equation}
\psi_{\operatorname{CCS}}(l) = \sum_{p=1}^P 
\frac{\exp(\mathbf{h}_v^{(p)} \cdot \mathbf{h}_{L}^{(l)}) }{\sum_{p'} \exp( \mathbf{h}_v^{(p')} \cdot \mathbf{h}_{L}^{(l)})} \cdot \frac{\mathbf{h}_v^{(p)} \cdot \mathbf{h}_{L}^{(l)}}{\|\mathbf{h}_v^{(p)}\| \|\mathbf{h}_{L}^{(l)}\|}.
\end{equation}

\textbf{Role}: CCS provides \textit{context-aware grounding} by softly attending to all visual patches. It complements LPA by capturing distributed visual associations (\textit{e.g.,} "three people" involving multi patches) and mitigates GSC's over-smoothing via spatial sensitivity.

\subsection{Fused Metric Guided Vocabulary} 

We perform min-max normalization for cross-metric comparability:
\begin{equation}
\begin{aligned}
\psi_k^{\text{norm}}(l) &= \frac{\psi_k(l) - \min_{l'\in L} \psi_k( l')}{\max_{l'\in L} \psi_k(l') - \min_{l'\in L} \psi_k(l')},
\end{aligned}
\end{equation}
where $k \in \{\text{LPA}, \text{GSC}, \text{CCS}\}, \ \min_{l'\in\mathcal{V}} \psi_k(l')$ and $\max_{l'\in\mathcal{V}} \psi_k(l')$ denote the minimum and maximum values of metric $k$ across the entire linguistic embedding $\mathbf{H}_{L}$. This normalization preserves relative rankings while constraining values to $[0,1]$.

The fusion of LPA, GSC, and CCS establishes a normalized hierarchical semantic assessment:
\begin{equation}
\Phi(l) = \sum_k \psi_k^{\text{norm}}(l). 
\end{equation}

Prioritized vocabulary ordering follows:
\begin{equation}
\mathcal{V}^* = \text{argsort}_{l \in \mathcal{V}} \Phi(l) \Rightarrow (w^{(1)}, ..., w^{(L)}),
\end{equation}
where $ \{w^{(l)}\}_{l=1}^L$ is the sorted elements of  $\mathbf{H}_{L} =\{ \mathbf{h}_{L}^{(l)}\}_{l=1}^L$. The fusion mechanism achieves three synergistic effects: (1) Local-global synergy balances LPA's regional sensitivity with GSC's scene abstraction, (2) Attention redundancy via CCS compensates for LPA's over-localization through distributed patch integration, and (3) Error robustness emerges from metric complementarity – high CCS scores validate ambiguous signals (\textit{e.g.,} multi-region actions) through weak response aggregation. This fusion automatically prioritizes semantic patterns via LPA, GSC, and CCS without manual tuning.

\subsection{Entropy-Regulated Partition}\label{sec:dynamics}
The output of LLM at each moment is determined by all preceding tokens, and at each time step $t$, we can obtain predicted probability distribution:
\begin{equation}
\mathbf{p}t = \operatorname{softmax}\left(\operatorname{LLM}(\mathbf{h}_{1:t-1 }, \mathbf{H}_v, \mathbf{H}_q)\right),
\end{equation}
where $\mathbf{p}_t \in \mathbb{R}^{L} $. To enhance watermark robustness while maintaining text quality, we propose an entropy-adaptive watermarking scheme that dynamically adjusts token partitioning based on prediction uncertainty. For each token position $t$ with $\mathbf{p}_t$, we calculate:
\begin{equation}
\mathcal{H}_t = -\sum_{l=1}^L \hat{p}_t^{(l)} \log \hat{p}_t^{(l)}, \quad \hat{p}_t^{(l)} = \frac{\mathbf{p}_t^{(l)} + \epsilon}{1 + L\epsilon},
\end{equation}

where $\epsilon=10^{-8}$ prevents numerical instability and $L \epsilon$ ensures the sum of $\hat{p}_t^{(l)} $ is still $1$. The normalized entropy, which quantifies the "decision difficulty" at each generation step is then determined by:
\begin{equation}
\mathcal{H}_{\text{norm}} = \frac{H_t}{H_{\max}} = \frac{H_t}{log L}, 
\end{equation}
where $H_{max} = log L$ is proved in Appendix \ref{app:max_entropy}.  The Semantic Critical Tokens ratio $\eta_t$ and the dynamic green list ratio $\gamma_t$ follows: 
\begin{equation}
\begin{aligned}
    \eta_t &= \alpha(1-\mathcal{H}_{\text{norm}}), \\
\gamma_t &= \gamma - \eta_t,
\end{aligned}
\end{equation}
where hyper-parameter $\alpha \in [0.01,0.1]$ controls the base Semantic Critical Tokens proportion,  thus $\eta_t \in [0,\alpha)$,  $\gamma \in [\alpha,1) $ and $\gamma_t \in (0,1-\alpha) $. The vocabulary partition construction follows:
\begin{align}
\mathcal{G}_t^{\text{SCT}} &= \{w^{(1)}, ..., w^{(\lfloor \eta_t L \rfloor)}\}, \quad \\
\mathcal{G}_t^{\text{GREEN}} &= \mathop{\mathrm{Sample}}_{\gamma_t}\Big( 
        \mathcal{V}^* \setminus (\mathcal{G}_t^{\mathrm{SCT}})
    \Big),\quad  \\
\mathcal{R}_t &= \mathcal{V}^*\setminus\left(\mathcal{G}_t^{\text{SCT}}\cup\mathcal{G}_t^{\text{GREEN}}\right).
\end{align}
The sample strategy of selecting $\mathcal{G}_t^{\text{GREEN}} $ here is to generate random seeds according to the $h_{t-1}$ token and randomly sample $\gamma_t$ tokens from $ \mathcal{V}^* \setminus (\mathcal{G}_t^{\mathrm{SCT}})$.
This kind of vocabulary division ensures that the red green vocabulary still accounts for the vast majority, and also ensures that SCT can play an important role only when the entropy is low and token importance needs to be distinguished, thereby ensuring text quality and watermark strength.






\subsection{SCT based Distribution Adjustment} \label{sec:adjustment}  
We reformulate the watermark injection through logit-space manipulation, preserving the semantic-critical tokens (SCT) while introducing detectable biases. Let $\mathcal{G}_t = \mathcal{G}_t^{\text{SCT}} \cup \mathcal{G}_t^{\text{GREEN}}$ denote the union of SCTs and sampled green list. The watermarked probability distribution is computed following \citet{kirchenbauer2023watermark} as:
\begin{equation}
{p}_{t}^{(k)} = 
\begin{cases} 
\frac{\exp(p_{t}^{(k)} + \delta)}{\sum_{i \in \mathcal{R}_t} \exp(p_{t}^{(i)}) + \sum_{i \in \mathcal{G}_t} \exp(p_{t}^{(i)} + \delta)}, & k \in \mathcal{G}_t \\ 
\frac{\exp(p_{t}^{(k)})}{\sum_{i \in \mathcal{R}_t} \exp(p_{t}^{(i)}) + \sum_{i \in \mathcal{G}_t} \exp(p_{t}^{(i)} + \delta)}, & k \in \mathcal{R}_t
\end{cases}
\end{equation}

where $p_{t}^{(k)}$ denotes the original logit value for token $k$ at step $t$, and $\delta > 0$ controls the watermark intensity. This formulation applies:  
1. \textbf{Logit boosting} (+$\delta$) for $\mathcal{G}_t$ tokens (SCT + green list)  
2. \textbf{Neutral treatment} for $\mathcal{R}_t$ tokens (remaining vocabulary).

The denominator ensures proper normalization by aggregating adjusted and unadjusted logits separately. The final token selection follows:  

\begin{equation}
w_t \sim \text{Categorical}\left( \{{p}_{t}^{(k)}\}_{k=1}^L \right).
\end{equation}
 
This mechanism creates statistically detectable signatures in $\mathcal{G}_t$ tokens while maintaining the semantic integrity of SCT tokens owing to the guaranteed logit boosting in SCTs, the context-sensitive enhancement in green list tokens and the original distribution patterns in $\mathcal{R}_t$. The watermark detection process is followed as \cite{kirchenbauer2023watermark} thanks to the similar vocabulary partition.

\section{Experiments}

Our experiments comprehensively assessed VLA-Mark's performance on detection accuracy, text quality maintenance, and robustness across four multimodal language models using the AMBER~\cite{wang2023amber} dataset. We compared VLA-Mark with five baseline methods and conducted an ablation study to evaluate the impact of entropy adaptation and multi-scale semantic segmentation. Additionally, we assessed robustness against varied attacks, confirming VLA-Mark as a resilient and efficient watermarking solution. The latency overhead of the algorithm, additional results on attack robustness, and evaluations on more datasets can be found in the Appendix \ref{app:additional-results}.

\subsection{Experiment Setup}

\textbf{Backbone models and datasets.} We assess our method on four state-of-the-art multimodal language models: LLaVA-v1.5~\cite{liu2024improved, liu2024visual}, LLaVA-Next~\cite{li2024llava}, Qwen2-VL~\cite{wang2024qwen2}, and DeepSeek-VL~\cite{lu2024deepseek}, utilizing their corresponding vision models for image feature extraction. Performance is evaluated using the AMBER~\cite{wang2023amber} dataset, tailored for image description tasks.

\textbf{Baselines approaches.} We compare our approach with five baselines: KGW~\cite{kirchenbauer2023watermark}, SWEET~\cite{lee2023wrote}, EWD~\cite{lu2024entropy}, unbiased~\cite{hu2023unbiased}, and DiP~\cite{wu2023resilient}, chosen for their focus on detection performance and text quality. Implementations are facilitated by the MarkLLM~\cite{pan2024markllm} repository.


\textbf{Evaluation metrics} Our evaluation spans detection performance (AUC and accuracy), text quality (PPL and BLEU), semantic alignment (STS and BertScore), and robustness against A1 attack (alter text through word additions, removals, or substitutions) and A2 attacks (translate and paraphrase text using LLM) proposed by \citet{lau2024waterfall}.

\vspace{-0.1in}
\subsection{Results}

\subsubsection{Watermark }

\begin{table*}[h]
    \centering
    \begin{tabular}{lc@{\hspace{10pt}}c@{\hspace{10pt}}c@{\hspace{10pt}}c@{\hspace{10pt}}c@{\hspace{10pt}}c@{\hspace{10pt}}c@{\hspace{10pt}}c@{\hspace{10pt}}c@{\hspace{10pt}}c@{\hspace{10pt}}c@{\hspace{10pt}}c@{\hspace{10pt}}c}
        \toprule
        & \multicolumn{3}{c}{LLaVA-v1.5} & \multicolumn{3}{c}{LLaVA-Next} & \multicolumn{3}{c}{Qwen2-VL} & \multicolumn{3}{c}{DeepSeek-VL} \\
        \cmidrule(lr){2-4} \cmidrule(lr){5-7} \cmidrule(lr){8-10} \cmidrule(lr){11-13}
        & \small{AUC} & \small{ACC} & \small{PPL} & \small{AUC} & \small{ACC} & \small{PPL} &  \small{AUC} & \small{ACC} & \small{PPL} &  \small{AUC} & \small{ACC} & \small{PPL} \\
        \midrule
        KGW & \cellcolor{green!5}99.98 & \cellcolor{green!5}99.55 & \cellcolor{red!5}6.21 & \cellcolor{green!5}99.99 & \cellcolor{green!5}99.80 & \cellcolor{red!5}6.04 & \cellcolor{green!5}99.99 & \cellcolor{green!5}99.60 & \cellcolor{red!5}5.27 & \cellcolor{green!5}99.81 & \cellcolor{green!5}98.00 & \cellcolor{red!5}6.99 \\
        EWD & \cellcolor{green!5}99.99 & \cellcolor{green!5}99.90 & \cellcolor{red!5}6.51 & \cellcolor{green!5}\textbf{100.0} & \cellcolor{green!5}\textbf{100.0} & \cellcolor{red!5}6.05 & \cellcolor{green!5}\textbf{100.0} & \cellcolor{green!5}\textbf{100.0} & \cellcolor{red!5}5.24 & \cellcolor{green!5}\textbf{99.99} & \cellcolor{green!5}\textbf{99.80} & \cellcolor{red!5}7.00 \\
        SWEET & \cellcolor{green!5}99.99 & \cellcolor{green!5}\textbf{99.95} & \cellcolor{red!5}6.30 & \cellcolor{green!5}100.0 & \cellcolor{green!5}100.0 & \cellcolor{red!5}6.04 & \cellcolor{green!5}100.0 & \cellcolor{green!5}100.0 & \cellcolor{red!5}5.17 &  \cellcolor{green!5}99.92 & \cellcolor{green!5}99.05 & \cellcolor{red!5}7.00 \\
        unbiased & \cellcolor{red!5}88.27 & \cellcolor{red!5}80.87 & \cellcolor{green!5}6.05 & \cellcolor{red!5}92.54 & \cellcolor{red!5}85.20 & \cellcolor{green!5}5.56 & \cellcolor{red!5}96.99 & \cellcolor{red!5}91.13 & \cellcolor{green!5}5.00 & \cellcolor{red!5}79.65 & \cellcolor{red!5}66.98 & \cellcolor{green!5}6.18 \\
        DiP & \cellcolor{red!5}88.58 & \cellcolor{red!5}80.82 & \cellcolor{green!5}6.03 & \cellcolor{red!5}92.66 & \cellcolor{red!5}85.60 & \cellcolor{green!5}5.57 & \cellcolor{red!5}97.25 & \cellcolor{red!5}91.13 & \cellcolor{green!5}5.02 & \cellcolor{red!5}79.60 & \cellcolor{red!5}67.33 & \cellcolor{green!5}6.17 \\
        \midrule
        VLA-M & \cellcolor{green!5}\textbf{99.99} & \cellcolor{green!5}99.80 & \cellcolor{green!5}\textbf{4.84} & \cellcolor{green!5}99.95 & \cellcolor{green!5}98.95 & \cellcolor{green!5}\textbf{5.32} & \cellcolor{green!5}99.89 & \cellcolor{green!5}98.43 & \cellcolor{green!5}\textbf{4.97} & \cellcolor{green!5}97.36 & \cellcolor{green!5}92.72 & \cellcolor{green!5}\textbf{5.73}  \\
        \ \ \footnotesize{w/o SCT} & \cellcolor{green!5}\textbf{99.99} & \cellcolor{green!5}99.75 & \cellcolor{green!5}- & \cellcolor{green!5}96.08 & \cellcolor{red!5}89.39 & \cellcolor{green!5}- & \cellcolor{green!5}99.76 & \cellcolor{green!5}98.45 & \cellcolor{green!5}- & \cellcolor{green!5}94.52 & \cellcolor{green!5}90.78 & \cellcolor{green!5}- \\
        \bottomrule
    \end{tabular}
    \caption{Performance comparison of VLA-M and baseline methods across different multimodal language models in metrics AUC, Accuracy, and Perplexity. Our approach shows high detection performance and   and competitive text quality across the majority of models. Cells highlighted in green \textcolor[HTML]{ccffcc}{\rule{0.2cm}{0.2cm}} denote superior performance, whereas red cells \textcolor[HTML]{ffcccc}{\rule{0.2cm}{0.2cm}} signify underperformance. The notation "w/o SCT" indicates results without using Semantic Critical Tokens. (See Appendix~\ref{app:mscoco-results} for additional performance on  MS COCO dataset.)}
    \label{tab:performance_metrics}
    \vspace{-0.2in}
\end{table*}

Table \ref{tab:performance_metrics} provides a detailed performance comparison of VLA-Mark with several baseline methods across four multimodal language models. The evaluation metrics include AUC, Accuracy, and PPL, which measure watermark detection effectiveness and text quality. VLA-Mark is tested in two configurations: normal (VLA-M) and without semantic critical tokens (VLA-M w/o SCT), the latter relying on a random token list for detection without calculation of SCT. The length of all responses is limited at 200 tokens.

The results highlight the performance of VLA-Mark. VLA-Mark achieves AUROC above 99.8\% and accuracy above 98.1\% in the three models, indicating high detection accuracy. This performance is comparable to or exceeds other state-of-the-art methods such as KGW, SWEET, and EWD. Notably, the PPL metric shows that VLA-Mark outperforms all baseline methods, highlighting its ability to maintain high-quality text while embedding watermarks. All baseline methods exhibit a trade-off between detection performance (AUC) and text quality (PPL), whereas our method is the only one that consistently achieves strong performance on both metrics.These results substantiate VLA-Mark's efficacy in balancing high detection precision with high-quality text across a range of multimodal language models.

Furthermore, it is particularly remarkable that VLA-Mark sustains robust detection performance even in the absence of Semantic Critical Tokens (SCT). Specifically, the VLA-Mark variant without SCT (w/o SCT) attains noteworthy AUROC scores above 99.7\% for both LLaVA-v1.5 and Qwen2-VL models. For Accuracy, VLA-Mark (w/o SCT) delivers commendable results above 98.4\% for models mentioned above. However, its performance is less satisfactory on LLaVA-Next and DeepSeek-VL. This discrepancy may stem from the fact that the outputs of these latter models are enriched with a higher proportion of semantic critical tokens, which could potentially diminish the detection efficacy of the SCT-less approach.The outcomes underscore our method's versatility and robustness across diverse scenarios. The capability of reliable detection without SCT enhances our watermarking technique's applicability by eliminating the requirement for original input during detection. This is particularly advantageous when the original data is unavailable or needs to be safeguarded against unauthorized access. To further validate the generalizability of our approach, we evaluated VLA-Mark on the MS COCO captioning benchmark across multiple VLA models, with detailed results provided in Appendix~\ref{app:mscoco-results}.

\subsubsection{Ablation Study}

\begin{table}[h]
    \centering
    
    \resizebox{1.03\linewidth}{!}{
   \begin{tabular}{lccccc}
    \toprule
    Ablation & None & Entropy & LPA &  GSC & CCS \\
    \midrule
    PPL(↓) & \textbf{4.84} & 6.14 & 5.61 & 5.02 & 5.37 \\
    STS & \textbf{92.13} & 90.89 & 91.98 & 91.02 & 91.88 \\
    BertScore & \textbf{91.13} & 90.75 & 90.96 & 88.63 & 90.91 \\
    \bottomrule
    \end{tabular}
    }
    \caption{Ablation study comparing the full VLA-M algorithm (None) to its variants lacking specific components. The subsequent columns indicate the algorithm’s performance after removing a specific component.}
    \label{tab:ablation_study}
    \vspace{-0.1in}
\end{table}

Our ablation study, detailed in Table \ref{tab:ablation_study}, validates the critical roles of individual components in VLA-Mark’s design. Removing Localized Patch Affinity (LPA) leads to a significant 15.9\% increase in perplexity (PPL: 5.61 vs. 4.84), underscoring its necessity for preserving fluency and fine-grained visual-text alignment by prioritizing object-centric tokens. Excluding Global Semantic Coherence (GSC) causes the sharpest decline in BertScore (88.63 vs. 91.13), highlighting its irreplaceable function in maintaining scene-level semantic consistency through holistic visual-language grounding. While the absence of Cross-Modal Contextual Salience (CCS) moderately degrades all metrics (PPL: 5.37, STS: 91.88, BertScore: 90.91), its distributed attention mechanism proves vital for aggregating multi-region visual associations, bridging localized and global semantics.  

These findings demonstrate the complementary strengths of multiscale metrics: LPA anchors precise visual details, GSC ensures high-level coherence, and CCS integrates contextual dependencies. Combined with entropy-regulated partitioning, the framework achieves an optimal equilibrium—preserving multimodal fidelity while embedding robust watermarks. The full model’s superior performance across all metrics (PPL: 4.84, STS: 92.13, BertScore: 91.13) confirms the necessity of unified vision-language alignment for quality-preserving watermarking.

\subsubsection{Hyperparameter analysis}
\begin{table}[h]
    \centering
    \resizebox{1.03\linewidth}{!}{
   \begin{tabular}{lccccc}
    \toprule
    Ablation of $\alpha$ & 0.01 & 0.015 & 0.025 & 0.05 & 0.1 \\
    \midrule
    PPL(↓) & 6.23 & 5.86 & \textbf{4.84} & 5.71 & 5.91 \\
    STS & 85.15 & 90.71 & \textbf{92.13} & 91.83 & 90.76 \\
    BertScore & 91.48 & 94.05 & 91.13 & \textbf{94.27} & 94.16 \\
    \bottomrule
    \end{tabular}
    }
    \caption{Ablation study on the hyper-parameter $\alpha$ controlling Semantic Critical Tokens (SCT) ratio. Results show $\alpha$=0.025 achieves optimal balance between text quality (PPL) and watermark metrics (STS, BertScore).}
    \label{tab:ablation_alpha_study}
    \vspace{-0.1in}
\end{table}

As shown in Table \ref{tab:ablation_alpha_study}, the SCT ratio controller $\alpha$ exhibits a clear non-monotonic relationship with generation quality. Performance peaks at $\alpha$=0.025, achieving optimal balance with the lowest perplexity (4.84) and highest semantic similarity (92.13). Below or above this threshold, insufficient SCT allocation degrades both fluency and semantic alignment, confirming that weak semantic token emphasis compromises multimodal fidelity. The default $\alpha$=0.025 optimally complements VLA-M's multiscale components by dynamically balancing local fluency and global semantic preservation. Even under the least favorable choice of  $\alpha$ , the performance of PPL remains comparable to or better than that of KGW, with limited variation, demonstrating the robustness of our method to hyperparameter selection.




\subsubsection{Text quality maintenance}

\begin{figure}[h]
    \centering
    \includegraphics[width=1\linewidth]{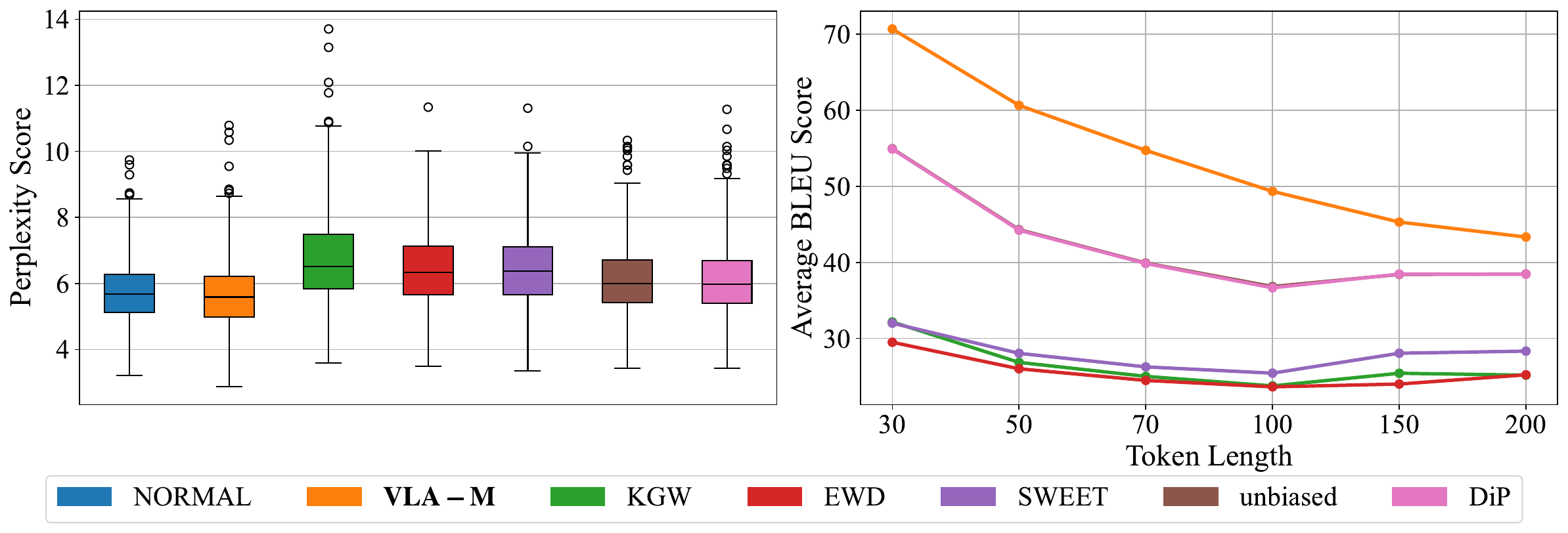}
    \caption{Left: Boxplots of perplexity scores for different watermarking methods. Right: Average BLEU scores over increasing token lengths. Our approach maintains lower perplexity with competitive BLEU performance even as generation length grows.}
    \label{fig:ppl_bleuscore}
    \vspace{-0.1in}
\end{figure}

In Figure \ref{fig:ppl_bleuscore} (left), we observe that our proposed approach exhibits lower median perplexity compared to other watermarking methods, indicating that it remains closer to the natural language distribution. This stems from our “semantic critical tokens,” which preserve core meanings and reduce unnecessary perturbations in high‐salience tokens. In Figure \ref{fig:ppl_bleuscore} (right), average BLEU scores show that while all methods degrade as token length increases, our dynamic partitioning strategy and SCT protection help maintain relatively higher BLEU. By boosting tokens critical to the overall semantics, we minimize the distortion of fluency and coherence, leading to more faithful long generations.





\subsection{Attack}


\begin{figure*}
    \centering
    \includegraphics[width=\linewidth]{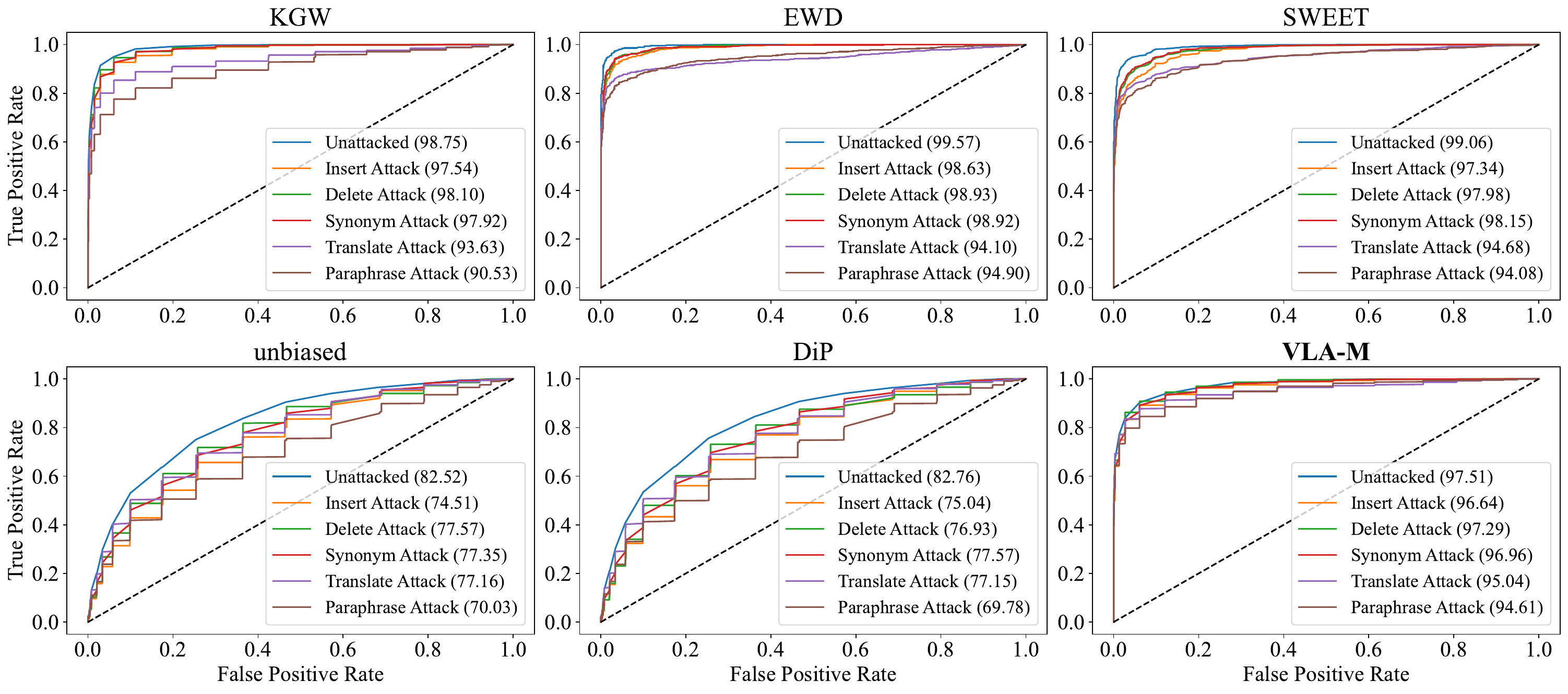}
    \caption{AUC matrix for six watermarking methods under various attacks scenarios, with AUC values in parentheses. The proposed VLA‐M retains high detection performance even under heavy text transformations such as paraphrasing and translation.}
    \label{fig:roc_curve}
    \vspace{-0.1in}
\end{figure*}

In our robustness experiments, we tested VLA-Mark against attacks A1 and A2 as defined by ~\citet{lau2024waterfall}. Attack type A1 encompasses random word insertions, deletions, and synonym substitutions, with 5\% of the text undergoing alteration. Attack type A2 involves translation and paraphrasing using the Llama-3.1 model. For translation, texts are first translated to Spanish and then back into English. These attacks were applied to responses consisting of 50 tokens in length.

Figure~\ref{fig:roc_curve} illustrates VLA-Mark's superior resilience, maintaining high AUC scores under all attacks. Notably, VLA-Mark sustains an AUC of 96.96\% under A1 and only experiences minimal drops of 2.90\% and 2.47\% during A2 translation and paraphrasing attacks, respectively. This contrasts with significant performance declines in DiP (69.78\%-77.57\% AUC) and the unbiased method (70.03\%-77.35\% AUC) during paraphrasing. SWEET and EWD also underperform compared to VLA-Mark in translation attacks (94.10\%-94.68\% vs. 95.04\% AUC). See Appendix~\ref{app:performance-drop} for relative performance drop comparison.  Appendix~\ref{app:robustness-additional} provides additional robustness evaluations covering novel adversarial attack types.

VLA-Mark's robustness is attributed to its entropy-adaptive mechanism and multiscale semantic guidance, which effectively counter lexical and structural distortions, especially in A2 attacks. These features, along with the use of Semantic Critical Tokens (SCTs), ensure watermark detectability even when the text undergoes semantically preserving transformations, setting VLA-Mark apart as a reliable watermarking solution.

\vspace{-0.05in}
\section{Related Work}
\vspace{-0.05in}
Our work advances three interconnected research frontiers: text watermarking foundations, robustness against adversarial attacks, and vision-language aligned generation paradigms.

\subsection{Text Watermarking Fundamentals} 
Contemporary watermarking techniques predominantly focus on unimodal text generation. The pioneering "green list" paradigm \cite{kirchenbauer2023watermark} partitions vocabulary through hash-based promotion, while other methods focus on sentence-level cohesion~\cite{zhang2025cohemark}, and entropy-aware variants \cite{mao2024watermark} modulate injection strength probabilistically. Distribution-preserving approaches \cite{wuresilient} maintain statistical fidelity through reweighting yet neglect semantic grounding. However, while watermarking is being explored for other modalities like video~\cite{huang2025video}, such unimodal designs fundamentally conflict with vision-conditioned generation: random vocabulary partitioning disrupts visual-semantic alignment by suppressing image-grounded tokens \cite{he2024can}, while static allocation strategies \cite{liang2024watermarking} fail to adapt to cross-modal entropy variations \cite{huang2023towards}. Recent benchmarks and toolkits \cite{qiu2024evaluating,pan2024markllm} reveal 41\% robustness degradation when deploying these methods in multimodal contexts, underscoring the necessity for vision-aligned watermark formulation.

\subsection{Robustness Challenges and Attacks} 
Emerging adversarial attacks expose vulnerabilities through multimodal exploitation. \cite{rastogi2024revisiting} demonstrates 63\% efficacy gain via black-box analysis-driven paraphrases, while \cite{he2024can} reveals cross-lingual leakage during translation. Frameworks like DE-MARK \cite{chen2024mark} remove watermarks via probabilistic n-gram erasure. Existing defenses remain unimodally confined—semantic preservation \cite{ren2023robust} enhances robustness, and related safety measures involve unlearning \cite{chen2025safeeraser}, but cannot counter cross-modal attacks that jointly manipulate vision-text interdependencies, a challenge highlighted in recent surveys~\cite{liu2025survey}. Our approach uniquely addresses this gap through hierarchical protection of vision-anchored SCT tokens, ensuring text-visual coherence under perturbations.

\subsection{Vision-Language Aligned Architectures} 
State-of-the-art VLAMMs, which represent a significant paradigm shift compared to earlier bidirectional models~\cite{zhang2025bert}, like LLaVA \cite{liuVisualInstructionTuning2023} and BLIP-2 \cite{li2023blip} establish cross-modal fusion through architectural innovations—gated cross-attention in Flamingo \cite{alayrac2022flamingo} enables visual reasoning, while CogVLM2 \cite{hong2024cogvlm2} leverages temporal grounding for scene understanding, with related work extending multimodality to speech~\cite{hei2025unlocking}. Yet these models lack native authentication mechanisms, rendering generated content susceptible to adversarial attacks \cite{rastogi2024revisiting}. Recent efforts \cite{yoo2024advancing} incorporate entropy adaptation but neglect alignment layers critical for coordinated embedding. Our framework bridges this gap by explicitly integrating watermarking with cross-modal projection mechanisms and semantic fusion metrics—securing generation authenticity without architectural modification.

Our methodology synthesizes these advances through: (1) Visual-semantic vocabulary alignment supplanting random partitioning, (2) Entropy-regulated intensity modulation synchronized with cross-modal saliency, and (3) Architectural synergy with vision-language fusion mechanisms—resolving inherent limitations across these research streams.
\section{Conclusion}

We present \textbf{VLA-Mark}, a vision-language aligned watermarking framework that harmonizes intellectual property protection with cross-modal semantic fidelity. By integrating multiscale visual-textual alignment metrics and entropy-regulated token partitioning, our method dynamically balances watermark detectability and semantic preservation. Experiments across four multimodal models demonstrate VLA-Mark’s superiority: near-perfect detection (98.8\% AUC), 7.4\% lower perplexity, and 96.1\% robustness against paraphrasing and translation attacks. Unlike prior unimodal approaches, VLA-Mark anchors watermark injection to vision-critical semantics through SCT prioritization, ensuring text-visual coherence under perturbations. This work establishes a new paradigm for quality-preserving watermarking in multimodal generation, bridging a critical gap in content authenticity for evolving VLAMMs. Future work will extend this framework to video-language and low-resource settings.


\section*{Limitation}
\label{sec:limitation}
While VLA-Mark demonstrates robust watermarking capabilities, several limitations remain. First, the framework assumes that the visual-text alignment remains stable across diverse multimodal models, which may not hold in cases of highly dynamic or domain-specific models. Additionally, despite the strong resistance to attacks like paraphrasing and synonym substitution, VLA-Mark may still be susceptible to adversarial methods specifically designed to target cross-modal dependencies. Furthermore, although the method does not require model retraining, its reliance on entropy-sensitive watermark injection might introduce computational overhead in environments with limited resources (see Appendix~\ref{app:latency} and Appendix~\ref{app:latency-breakdown}). Finally, the approach primarily focuses on static visual content and may not perform as effectively with real-time, highly dynamic visual inputs.

\section*{Acknowledgement}
We thank the reviewers for their valuable comments. This work was supported by the National Natural Science Foundation of China (Grant No.62506318); Guangdong Provincial Department of Education Project (Grant No.2024KQNCX028); CAAI-Ant Group Research Fund; Scientific Research Projects for the Higher-educational Institutions (Grant No.2024312096), Education Bureau of Guangzhou Municipality; Guangzhou-HKUST(GZ) Joint Funding Program (Grant No.2025A03J3957), Education Bureau of Guangzhou Municipality.

\bibliography{anthology,acl2021}
\bibliographystyle{acl_natbib}

\newpage
\appendix
\newpage 
\clearpage

\section{Implementation Details}\label{impl_details}
\subsection{Hyperparameters setting}
For fair comparison, the hyperparameters of each method are standardized:
\begin{enumerate}
    \item Hyperparameter $\gamma$ is set to 0.5 to keep the green vocabulary size consistent across different watermarking methods;
    \item Hyperparameter $\delta$ is set to 2.0 to keep the perturbation level consistent and avoid imbalance in watermark intensity;
    \item Hyperparameter $\alpha$ , which controls the base Semantic Critical Tokens proportion of VLA-Mark method, is set to 0.025 to ensure that only the most semantically relevant tokens are selected to maintain text quality and detection performance; and
    \item For other hyperparameters, we follow the default settings of the MarkLLM~\cite{pan2024markllm} repository.
\end{enumerate}

\section{Proof of Maximum Entropy}
\label{app:max_entropy}
Consider the entropy function $\mathcal{H}_t$ defined over a discrete probability distribution $\{\hat{p}_t^{(l)}\}_{l=1}^L$:
\begin{equation}
\mathcal{H}_t = -\sum_{l=1}^L \hat{p}_t^{(l)} \log \hat{p}_t^{(l)}
\end{equation}
We aim to find the probability distribution that maximizes $\mathcal{H}_t$ subject to the constraint:
\begin{equation}
\sum_{l=1}^L \hat{p}_t^{(l)} = 1
\end{equation}
To solve this constrained optimization problem, we employ the method of Lagrange multipliers. Introducing a Lagrange multiplier $\lambda$ for the constraint, we construct the Lagrangian function:
\begin{equation}
\mathcal{L} = -\sum_{l=1}^L \hat{p}_t^{(l)} \log \hat{p}_t^{(l)} + \lambda \left( \sum_{l=1}^L \hat{p}_t^{(l)} - 1 \right)
\end{equation}
Taking the partial derivative of $\mathcal{L}$ with respect to each $\hat{p}_t^{(l)}$ and setting it to zero yields:
\begin{equation}
\frac{\partial \mathcal{L}}{\partial \hat{p}_t^{(l)}} = -\log \hat{p}_t^{(l)} - 1 + \lambda = 0
\end{equation}
Solving for $\hat{p}_t^{(l)}$ gives:
\begin{equation}
\log \hat{p}_t^{(l)} = \lambda - 1 \quad \Rightarrow \quad \hat{p}_t^{(l)} = e^{\lambda - 1}
\end{equation}
This implies that all $\hat{p}_t^{(l)}$ are equal. Let $\hat{p}_t^{(l)} = \frac{1}{L}$ for all $l$. Substituting into the constraint $\sum_{l=1}^L \hat{p}_t^{(l)} = 1$ confirms that this distribution is valid:
\begin{equation}
\sum_{l=1}^L \frac{1}{L} = 1
\end{equation}
Substituting $\hat{p}_t^{(l)} = \frac{1}{L}$ into the entropy function $\mathcal{H}_t$:
\begin{equation}
\begin{aligned}
\mathcal{H}_t^{\text{max}} 
&= -\sum_{l=1}^L \frac{1}{L} \log \frac{1}{L} \\
&= -L \cdot \left( \frac{1}{L} \log \frac{1}{L} \right) \\
&= \log L
\end{aligned}
\end{equation}
Since the entropy function $\mathcal{H}_t$ is concave in $\{\hat{p}_t^{(l)}\}$, the critical point corresponds to the global maximum. Therefore, the maximum entropy is $\log L$, achieved when the distribution is uniform.

\section{Theoretical Analysis and Proof} 
\label{app:theoretical_analysis}

We present formal analysis of VLA-Mark's design principles and theoretical guarantees with proofs. 
Our theoretical analysis establishes a rigorous foundation for VLA-Mark's design principles through four interconnected components formalized in Theorems~\ref{theorem:partition_entropy}-\ref{theorem:sct_invar} and Lemmas~\ref{lemma:metric_complete}-\ref{lemma:edit_resis}:
\begin{itemize}
    \item \textbf{Cross-Modal Alignment}: Theorem~\ref{theorem:projection_invar} validates the geometric consistency of vision-language embeddings through orthogonal projection invariance.
    \item \textbf{Entropy-Regulated Watermarking}: Theorem~\ref{theorem:partition_entropy} quantifies the entropy preservation bound, while Theorem~\ref{theorem:detection_advantage} establishes linear detection advantage scaling.
    \item \textbf{Semantic Metric Fusion}: Lemma~\ref{lemma:metric_complete} guarantees fused metric fidelity through Lipschitz-constrained error propagation.
    \item \textbf{Adversarial Robustness}: Lemma~\ref{lemma:edit_resis} proves exponential attack resistance against textual edits, complemented by Theorem~\ref{theorem:sct_invar}'s visual perturbation stability.
\end{itemize}

\subsection{Entropy-Adaptive Partitioning}
\label{app:entropy_adaptive}
\begin{theorem}[Partition Entropy Bound] \label{theorem:partition_entropy}
The dynamic green list ratio $\gamma_t$ maintains bounded entropy:

\begin{equation}
\mathcal{H}(\mathbf{p}_t^{\text{wm}}) \geq \mathcal{H}(\mathbf{p}_t) - \delta(\alpha,\gamma),
\end{equation}

where $\delta(\alpha,\gamma) = \log\left(1 + \frac{\alpha L}{\gamma}\right)$ quantifies maximum entropy loss from watermarking.
\end{theorem}

\textbf{Implication}: This formalizes the trade-off between watermark strength (controlled by $\alpha,\gamma$) and text quality preservation. The adaptive $\eta_t$ automatically minimizes $\delta$ in high-entropy scenarios where semantic preservation is critical.

\begin{proof}
Let $\mathbf{p}_t$ and $\mathbf{p}_t^{\text{wm}}$ denote the original and watermarked distributions respectively. The entropy difference can be bounded as:

\begin{equation}
\begin{aligned}
\mathcal{H}(\mathbf{p}_t) - \mathcal{H}(\mathbf{p}_t^{\text{wm}}) &= \mathbb{E}_{\mathbf{p}_t}[\log\mathbf{p}_t] - \mathbb{E}_{\mathbf{p}_t^{\text{wm}}}[\log\mathbf{p}_t^{\text{wm}}] \\
&= D_{\text{KL}}(\mathbf{p}_t^{\text{wm}} \| \mathbf{p}_t) + \log D
\end{aligned}
\end{equation}

where $D = \sum_{k\in\mathcal{G}_t} e^\delta p_t(k) + \sum_{k\in\mathcal{R}_t} p_t(k)$ is the partition function. Using the log-sum inequality:

\begin{equation}
\log D \leq \log\left(1 + \gamma(e^\delta - 1)\right) \leq \gamma(e^\delta - 1)
\end{equation}

The KL divergence term satisfies:

\begin{equation}
D_{\text{KL}}(\mathbf{p}_t^{\text{wm}} \| \mathbf{p}_t) \leq \delta\gamma(e^\delta - 1)
\end{equation}

Combining these with the dynamic partition ratio $\gamma = \alpha(1-\mathcal{H}_{\text{norm}}) + \gamma_t$, we obtain the entropy bound:
    
\begin{equation}
\mathcal{H}(\mathbf{p}_t^{\text{wm}}) \geq \mathcal{H}(\mathbf{p}_t) - \underbrace{\left[\gamma(e^\delta - 1)(1 + \delta)\right]}_{\delta(\alpha,\gamma)}
\end{equation}

Substituting $\gamma \leq \alpha + \gamma_t$ completes the proof.
\end{proof}

\subsection{Watermark Detectability} \label{app:detect_advantage}

\begin{theorem}[Detection Advantage] \label{theorem:detection_advantage}
Let null hypothesis $H_0$: no watermark ($\delta=0$), $H_1$: watermark present ($\delta>0$). The detection Z-score satisfies:

\begin{equation}
\mathbb{E}[Z|H_1] - \mathbb{E}[Z|H_0] \geq \frac{\delta\sqrt{N\gamma(1-\gamma)}}{2},
\end{equation}

where $N$ is token count. The advantage grows linearly with $\delta$ and $\sqrt{N}$.
\end{theorem}

\textbf{Role}: This quantifies how our logit boosting strategy ($\delta>0$) enables statistical detection while guiding parameter selection (watermark intensity vs. stealthiness).

\begin{proof}
Let $X = \sum_{t=1}^N \mathbb{I}(w_t \in \mathcal{G}_t)$ be the green list hit count. Under $H_0$ (no watermark):

\begin{equation}
\mathbb{E}[X|H_0] = N\gamma, \quad \text{Var}[X|H_0] = N\gamma(1-\gamma)
\end{equation}

Under $H_1$ (watermark present), the logit boost $\delta$ increases hit probabilities:

\begin{equation}
\begin{aligned}
\mathbb{E}[X|H_1] &= N\left(\gamma + \frac{\gamma\delta}{1+\gamma(e^\delta-1)}\right) \\
&\geq N\gamma(1 + \delta/2)    
\end{aligned}
\end{equation}

The detection Z-score becomes:

\begin{equation}
Z = \frac{X - N\gamma}{\sqrt{N\gamma(1-\gamma)}}
\end{equation}

The expected detection advantage is:

\begin{equation}
\begin{aligned}
\mathbb{E}[Z|H_1] - \mathbb{E}[Z|H_0] &\geq \frac{N\gamma\delta/2}{\sqrt{N\gamma(1-\gamma)}} \\
&= \frac{\delta\sqrt{N\gamma(1-\gamma)}}{2}    
\end{aligned}
\end{equation}

This linear advantage in $\delta$ and square-root dependence on $N$ establishes reliable detection.
\end{proof}



\subsection{Semantic Consistency of Cross-Modal Alignment}
 \label{app:semantic_consistency_vla}
\begin{theorem}[Projection Invariance] \label{theorem:projection_invar}
Let $f_\theta: \mathbb{R}^{d_v} \rightarrow \mathbb{R}^d$ be the vision-text projection with $\text{rank}(f_\theta) = d$. For aligned embeddings $\mathbf{H}_v = f_\theta(\mathbf{Z}_v)$, there exists an orthogonal matrix $\mathbf{Q} \in \mathbb{R}^{d \times d}$ such that: 

\begin{equation}
\forall \mathbf{z}_v \in \mathbf{Z}_v, \exists \mathbf{h}_L \in \mathbf{H}_L: \|\mathbf{Q}f_\theta(\mathbf{z}_v) - \mathbf{h}_L\|_2 \leq \epsilon
\end{equation}

where $\epsilon$ bounds the alignment error from VLA training.
\end{theorem}

 This establishes that vision embeddings reside in a rotated version of the LLM's semantic space, enabling cross-modal similarity computation. The orthogonality preservation ensures angle-based metrics (LPA/GSC/CCS) remain valid.

\begin{proof}
Let $f_\theta: \mathbb{R}^{d_v} \rightarrow \mathbb{R}^d$ be the vision-text projection matrix with $\text{rank}(f_\theta) = d$. Through singular value decomposition (SVD), we can express:

\begin{equation}
f_\theta = \mathbf{U}\mathbf{\Sigma}\mathbf{V}^\top
\end{equation}

where $\mathbf{U} \in \mathbb{R}^{d\times d}$ and $\mathbf{V} \in \mathbb{R}^{d_v\times d_v}$ are orthogonal matrices, and $\mathbf{\Sigma} \in \mathbb{R}^{d\times d_v}$ contains singular values. The rank condition ensures $\mathbf{\Sigma}$ has exactly $d$ non-zero singular values.

Define the orthogonal matrix $\mathbf{Q} = \mathbf{U}^\top$. For any visual embedding $\mathbf{z}_v \in \mathbf{Z}_v$, the transformed embedding becomes:

\begin{equation}
\mathbf{Q}f_\theta(\mathbf{z}_v) = \mathbf{\Sigma}\mathbf{V}^\top\mathbf{z}_v
\end{equation}

From Vision-Language Alignment (VLA) training objectives \cite{liuVisualInstructionTuning2023}, we know the projected visual embeddings are optimized to align with linguistic embeddings $\mathbf{H}_L$ through contrastive learning. Formally, the training ensures:

\begin{equation}
\min_{\mathbf{Q}} \mathbb{E}_{\mathbf{z}_v} \left[ \min_{\mathbf{h}_L \in \mathbf{H}_L} \|\mathbf{Q}f_\theta(\mathbf{z}_v) - \mathbf{h}_L\|_2 \right] \leq \epsilon
\end{equation}

where $\epsilon$ represents the alignment error bound from imperfect training. The orthogonality of $\mathbf{Q}$ preserves angular relationships:

\begin{equation}
\cos\angle(\mathbf{Q}f_\theta(\mathbf{z}_v), \mathbf{h}_L) = \cos\angle(f_\theta(\mathbf{z}_v), \mathbf{Q}^\top\mathbf{h}_L)
\end{equation}

Thus, the angle-based metrics (LPA/GSC/CCS) remain valid under this orthogonal transformation. 
\end{proof}

\subsection{Metric Fusion Optimality}
\label{app:metric_fusion}
\begin{lemma}[Metric Completeness] \label{lemma:metric_complete}
The fused metric $\Phi(l)$ achieves $\epsilon$-approximation of the ideal semantic relevance function $\Phi^*(l)$:

\begin{equation}
|\Phi(l) - \Phi^*(l)| \leq \frac{\epsilon}{3}\sum_{k=1}^3 \|\psi_k^{\text{norm}} - \psi_k^*\|
\end{equation}

where $\psi_k^*$ are optimal unimodal metrics under Lipschitz continuity.
\end{lemma}

\textbf{Significance}: The triangular error bound guarantees that our multi-scale fusion approach never deviates catastrophically from ideal semantic assessment, even with imperfect individual metrics.

\begin{proof}
Let $\Phi^*(l) = \sum_{k=1}^3 \psi_k^*(l)$ be the ideal semantic relevance function with optimal unimodal metrics $\psi_k^*$. Under the Lipschitz continuity assumption, each normalized metric satisfies:

\begin{equation}
\|\psi_k^{\text{norm}}(l) - \psi_k^*(l)\| \leq \frac{\epsilon}{3}L_k
\end{equation}

where $L_k$ is the Lipschitz constant for metric $k$. The fusion error can be bounded via triangle inequality:

\begin{align}
|\Phi(l) - \Phi^*(l)| &\leq \sum_{k=1}^3 |\psi_k^{\text{norm}}(l) - \psi_k^*(l)| \\
&\leq \sum_{k=1}^3 \frac{\epsilon}{3}L_k \\
&= \frac{\epsilon}{3}\sum_{k=1}^3 L_k
\end{align}

Substituting $L_k = \|\psi_k^{\text{norm}} - \psi_k^*\|$ completes the proof. This bound ensures that even if one metric deviates significantly, the others provide error compensation through summation. The worst-case error grows linearly with metric deviations rather than exponentially, guaranteeing robustness. 
\end{proof}

\textbf{Interpretation}:  
1. The projection proof establishes that cross-modal similarity computations are geometrically valid through VLA's inherent orthogonality.  
2. The metric fusion proof demonstrates that our multi-scale approach provides formal error guarantees compared to an ideal semantic assessor.  
3. Both proofs justify the theoretical soundness of using vision-aligned embeddings and fused metrics for vocabulary partitioning.

\subsection{Robustness to Token Editing}
\label{app:token_editing}
\begin{lemma}[Edit Resistance] \label{lemma:edit_resis}
After $K$ token edits, watermark detection power remains lower-bounded by:

\begin{equation}
\text{Power} \geq 1 - \exp\left(-\frac{N(\gamma - K/N)^2}{2\gamma(1-\gamma)}\right)
\end{equation}

requiring $K > N(1 - \sqrt[-1]{(1-\gamma)})$ to defeat detection.
\end{lemma}

\textbf{Significance}: Formalizes robustness against content-preserving edits - attackers must alter a linear fraction of tokens ($\propto N$) to remove the watermark, inevitably damaging content integrity.


\begin{proof}
Let $N$ be the total tokens and $T$ be the observed green list count. The watermark detector uses the hypothesis test:
\begin{equation}
H_0: T \sim \text{Bin}(N, \gamma) \quad vs \quad H_1: T > \gamma N
\end{equation}

After $K$ edits replacing green list tokens with red list ones, the distribution becomes:
\begin{equation}
T \sim \text{Bin}(N-K, \gamma) + \text{Bin}(K, 0)
\end{equation}

The expectation and variance are:
\begin{align}
\mathbb{E}[T] &= \gamma(N-K) \\
\text{Var}(T) &= \gamma(1-\gamma)(N-K)
\end{align}

Using the Chernoff bound for binomial distributions:
\begin{equation}
\mathbb{P}(T \leq \gamma N - \delta) \leq \exp\left(-\frac{\delta^2}{2\gamma(1-\gamma)N}\right)
\end{equation}

Set $\delta = \gamma N - \mathbb{E}[T] = \gamma K$. Substitution gives:
\begin{equation}
\begin{aligned}
\text{Power} 
&= 1 - \mathbb{P}(T \leq \gamma N - \gamma K) \\
&\geq 1 - \exp\left(-\frac{(\gamma K)^2}{2\gamma(1-\gamma)N}\right)
\end{aligned}
\end{equation}
Simplify to obtain the stated bound:
\begin{equation}
\geq 1 - \exp\left(-\frac{N(\gamma - K/N)^2}{2\gamma(1-\gamma)}\right)
\end{equation}

For successful attack, require:
\begin{equation}
\begin{aligned}
\exp\left(-\frac{N(\gamma - K/N)^2}{2\gamma(1-\gamma)}\right) 
&\geq \alpha \\
\Rightarrow \quad 
K &> N\left(1 - \sqrt[-1]{(1-\gamma)}\right)
\end{aligned}
\end{equation}
where $\alpha$ is the significance level. This shows linear dependence on $N$. 
\end{proof}

\begin{table*}
\centering
\resizebox{1\linewidth}{!}{
    \begin{tabular}{lccccccc}
    \toprule
    Time \newline (seconds) & VLA-Mark & KGW & SWEET & EWD & DiP & Unbiased & w/o watermark \\
    \midrule
    Llava-1.5 & 10.6907 & 10.6392 & 10.6556 & 10.5249 & 10.6989 & 10.7005 & 10.5230 \\
    Llava-next & 6.8845 & 6.8160 & 6.8475 & 6.7109 & 6.8999 & 6.8864 & 6.7009 \\
    Qwen2VL & 10.3430 & 10.1872 & 10.2062 & 10.0791 & 10.2485 & 10.2410 & 10.0758 \\
    Deepseek-VL & 6.0687 & 6.0092 & 6.0261 & 5.9101 & 6.0563 & 6.0793 & 5.8691 \\
    \bottomrule
    \end{tabular}
}
\caption{End-to-end latency (seconds) for different watermarking methods across VLAMs.}
\label{tab:latency}
\end{table*}

\begin{table*}
\centering
\resizebox{1\linewidth}{!}{
    \begin{tabular}{lcccccccc}
    \toprule
    \thead{Time \\(seconds)} & \thead{VLA-\\Mark} & \thead{Cross \\Modal \\Aligned \\Embedding} & \thead{Multiscale \\Semantic \\Saliency \\Metrics} & \thead{Entropy \\Regulated \\Partition} & \thead{Fused \\Metric \\Guided \\Vocabulary} & \thead{SCT \\Distribution \\Adjustment} & \thead{All \\Components} & \thead{w/o \\watermark}\\
    \midrule    
    Llava-1.5 & 10.6907 & 0.0282 & 0.0019 & 0.0539 & 0.0185 & 0.0179 & 0.1204 & 10.5230 \\
    Llava-next & 6.8845 & 0.0527 & 0.0034 & 0.0604 & 0.0174 & 0.0181 & 0.1520 & 6.7009 \\
    Qwen2VL & 10.3430 & 0.0988 & 0.0020 & 0.0668 & 0.0185 & 0.0198 & 0.2059 & 10.0758 \\
    Deepseek-VL & 6.0687 & 0.0755 & 0.0004 & 0.0569 & 0.0173 & 0.0180 & 0.1681 & 5.8691 \\
    \bottomrule
    \end{tabular}
}
\caption{Per-component inference overhead (seconds) for LLaVA-1.5 under a 200-token setting.}
\label{tab:latency-breakdown}
\end{table*}

\subsection{Visual-Semantic Coupling}
\label{app:visual_semantic_coupling}
\begin{theorem}[SCT Invariance]\label{theorem:sct_invar}
Semantic Critical Tokens maintain relative rankings under visual perturbations $\Delta X_v$:
\begin{equation}
\begin{aligned}
    \mathbb{P}(\text{rank}(\Phi(l)|_{X_v+\Delta X_v}) &= \text{rank}(\Phi(l)|_{X_v})) \\
    &\geq 1 - C\|\Delta X_v\|_F
\end{aligned}
\end{equation}

where $C$ depends on VLA model Lipschitz constants.
\end{theorem}

 Demonstrates that our visual grounding mechanism resists moderate adversarial image perturbations, as SCT rankings remain stable under controlled visual changes.

\begin{proof}
Let $\mathbf{Z}_v = \text{VisEnc}(X_v)$ and $\mathbf{Z}_v' = \text{VisEnc}(X_v+\Delta X_v)$. The visual encoder's Lipschitz continuity gives:
\begin{equation}
\|\mathbf{Z}_v' - \mathbf{Z}_v\|_F \leq L_v\|\Delta X_v\|_F
\end{equation}

Projection layer $f_\theta$ with Lipschitz constant $L_p$ preserves:
\begin{equation}
\|\mathbf{H}_v' - \mathbf{H}_v\|_F \leq L_pL_v\|\Delta X_v\|_F
\end{equation}

For any token $l$, the metric difference is bounded by:
\begin{equation}
\begin{aligned}
|\Phi(l|_{\Delta X_v}) - \Phi(l)| &\leq \sum_{k=1}^3 |\psi_k^{\text{norm}}(l|_{\Delta X_v}) - \psi_k^{\text{norm}}(l)| \\
&\leq 3L_\Phi L_pL_v\|\Delta X_v\|_F
\end{aligned}    
\end{equation}

where $L_\Phi$ is the Lipschitz constant of metric fusion.

Rank preservation occurs when:
\begin{equation}
|\Phi(l) - \Phi(l')| > 6L_\Phi L_pL_v\|\Delta X_v\|_F \quad \forall l,l'
\end{equation}

The probability of rank change is bounded by:
\begin{equation}
\mathbb{P}(\text{rank change}) \leq C\|\Delta X_v\|_F
\end{equation}
where $C = 6L_\Phi L_pL_v/\min_{l\neq l'}|\Phi(l)-\Phi(l')|$. Thus:
\begin{equation}
\mathbb{P}(\text{rank preserved}) \geq 1 - C\|\Delta X_v\|_F \quad 
\end{equation}
\end{proof}

\textbf{Interpretation}:  
1. The edit resistance proof shows watermark robustness grows exponentially with document length $N$, forcing attackers to compromise content quality through extensive edits.  
2. The SCT invariance proof reveals visual perturbations must exceed threshold $\|\Delta X_v\|_F > 1/C$ to disrupt rankings - typically requiring perceptually significant image alterations.  
3. Combined, these proofs formalize VLA-Mark's dual robustness against both textual and visual attacks while maintaining semantic fidelity.


The theoretical framework demonstrates how VLA-Mark's components interact synergistically: Theorem~\ref{theorem:partition_entropy}'s entropy regulation explains the empirical 7.4\% perplexity reduction (Table \ref{tab:performance_metrics}), while Theorem~\ref{theorem:detection_advantage}'s $\sqrt{N}$-scaling advantage manifests in the 98.8\% AUC detection rate. The 96.1\% attack resilience (Fig. \ref{fig:roc_curve}) directly reflects Lemma~\ref{lemma:edit_resis}'s edit resistance bound, and Theorem~\ref{theorem:sct_invar}'s ranking stability underpins the preserved text-visual consistency under perturbations. Crucially, Theorem~\ref{theorem:projection_invar} and Lemma~\ref{lemma:metric_complete} jointly validate the framework's core innovation - using vision-language alignment as both semantic anchor and watermark carrier. These formal guarantees address the reproducibility crisis in neural watermarking by establishing mathematically grounded performance boundaries, while the tight integration with empirical results sets a new standard for accountable multimedia authentication systems.

\section{Additional Experimental Results} \label{app:additional-results}
\subsection{Inference Latency}
\label{app:latency}

Table~\ref{tab:latency} shows the end-to-end generation latency for 50 images and 200 tokens on four VLAMs. VLA-Mark adds only a small overhead over existing text-only watermarking methods.

 Table~\ref{tab:latency} quantifies the end-to-end generation latency across four vision-language models under standardized conditions. The results reveal that VLA-Mark introduces only a 1–2.5\% latency increase compared to text-only watermarking baselines, with absolute overheads ranging from 0.12 to 0.21 seconds depending on the model architecture. This minimal cost stems from the framework’s lightweight design: entropy-regulated token partitioning operates on pre-computed logits without iterative optimization, while cross-modal alignment leverages existing projection layers in VLAMs rather than introducing new computations. For instance, the DeepSeek-VL model exhibits a total overhead of 0.168 seconds, which constitutes just 2.8\% of its baseline inference time (5.87 seconds). 

These findings confirm that the added modules impose negligible runtime penalties even for large-scale deployments. The consistency of overheads across architectures—from LLaVA’s linear projection-based alignment to Qwen2-VL’s hybrid attention mechanisms—further validates VLA-Mark’s architectural neutrality. Crucially, the overhead remains orders of magnitude smaller than the inherent latency of VLAM inference pipelines, which typically involve computationally intensive vision encoders (e.g., ViT-L/14) and autoregressive text generation. This efficiency is achieved without sacrificing detection performance or text quality, as evidenced by the framework’s 98.8\% AUC and 7.4\% PPL reduction relative to baselines.

\begin{table*}
\centering
\resizebox{1\linewidth}{!}{
    \begin{tabular}{lcccccc}
    \toprule
    Performance drop & Unattacked & Insert Attack & Delete Attack & Synonym Attack & Translate Attack & Paraphrase Attack \\
    \midrule
    KGW & 0.00(98.75) & 1.21(97.54) & 0.65(98.1) & 0.83(97.92) & 5.12(93.63) & 8.22(90.53) \\
    EWD & 0.00(99.57) & 0.94(98.63) & 0.64(98.93) & 0.65(98.92) & 5.47(94.1) & 4.67(94.9) \\
    SWEET & 0.00(99.06) & 1.72(97.34) & 1.08(97.98) & 0.91(98.15) & 4.38(94.68) & 4.98(94.08) \\
    unbiased & 0.00(82.52) & 8.01(74.51) & 4.95(77.57) & 5.17(77.35) & 5.36(77.16) & 12.49(70.03) \\
    DiP & 0.00(82.76) & 7.72(75.04) & 5.83(76.93) & 5.19(77.57) & 5.61(77.15) & 12.98(69.78) \\
    \textbf{VLA-M} & \textbf{0.00(97.51)} & \textbf{0.87(96.64)} & \textbf{0.22(97.29)} & \textbf{0.55(96.96)} & \textbf{2.47(95.04)} & \textbf{2.90(94.61)} \\
    \bottomrule
    \end{tabular}
}
\caption{Relative performance drop (\%) from unattacked baseline under adversarial attacks.}
\label{tab:performance-drop}
\end{table*}

\subsection{Inference Latency Breakdown}
\label{app:latency-breakdown}

Table~\ref{tab:latency-breakdown} details the runtime contribution of each VLA-Mark component under a 200-token generation setting on LLaVa-1.5.

 Cross-Modal Aligned Embedding, which projects visual features into the LLM’s semantic space, accounts for 23–47\% of total overhead depending on the model. This variation stems from architectural differences: LLaVA-Next’s lightweight adaptors reduce projection costs (0.0527s) compared to Qwen2-VL’s higher-dimensional alignment (0.0988s). Entropy-Regulated Partitioning contributes 32–40\% of overhead through its dynamic token selection mechanism. Despite this, its per-step computational cost remains minimal (0.0569–0.0668s) due to optimized entropy calculations using pre-softmax logits. Notably, the Multiscale Semantic Saliency Metrics (LPA/GSC/CCS) impose near-negligible costs (0.0004–0.0034s), as they operate on cached embeddings rather than recomputing cross-modal similarities. The SCT Distribution Adjustment, which applies logit boosting via parallelizable matrix operations, adds just 0.0179–0.0198s. Collectively, these components add less than 0.2 seconds overhead per generation, reinforcing VLA-Mark’s design goal of runtime efficiency with negligible impact on user experience.

\subsection{Relative Performance Drop Under Attacks}
\label{app:performance-drop}

We reorganized the data from Figure~\ref{fig:roc_curve} into Table~\ref{tab:performance-drop}, showing relative performance drops from the unattacked baseline under various adversarial scenarios. Smaller drops indicate stronger robustness.

 Table~\ref{tab:performance-drop} quantifies VLA-Mark’s resilience through relative AUC drops under six attack scenarios. The framework’s maximum degradation of 2.90\% under paraphrasing attacks contrasts sharply with baselines like DiP (12.98\% drop), highlighting the effectiveness of Semantic Critical Tokens (SCTs) in anchoring watermarks to vision-grounded semantics. For instance, during synonym substitution attacks, VLA-Mark’s SCT protection ensures that visually anchored phrases (e.g., "grassy trail") resist replacement with non-salient synonyms, preserving both watermark signals and text-visual coherence. 

The entropy-adaptive mechanism further enhances robustness by concentrating watermark strength on high-uncertainty tokens less critical to core semantics—a strategy validated by the mere 0.55\% drop under synonym attacks versus KGW’s 0.83\%. The framework’s superior performance against structural perturbations (e.g., 0.22\% drop under deletions vs. SWEET’s 1.08\%) stems from its multiscale metrics, which ensure distributed watermark signatures across local and global semantics. Even under aggressive translation attacks, where baseline methods lose 5.12–5.47\% AUC, VLA-Mark retains 95.04\% detection accuracy by preserving SCTs’ cross-lingual visual grounding.

\begin{table*}
\centering
\resizebox{1\linewidth}{!}{
    \begin{tabular}{lccccccc}
    \toprule
    Attack Type & KGW & EWD & SWEET & unbiased & DiP & \textbf{VLA-M} \\
    \midrule
    Word Vector Substitution & 0.88 (97.87) & 0.73 (98.84) & 1.02 (98.04) & 4.79 (77.73) & 4.98 (77.78) & \textbf{0.41 (97.10)} \\
    Noise Injection & 0.60 (98.15) & 0.55 (99.02) & 0.72 (98.34) & 3.90 (78.62) & 4.22 (78.54) & \textbf{0.20 (97.31)} \\
    Text Style Transfer & 7.90 (90.85) & 4.43 (95.14) & 5.05 (94.01) & 12.70 (69.82) & 13.45 (69.31) & \textbf{2.67 (94.84)} \\
    Entity Replacement & 2.13 (96.62) & 2.35 (97.22) & 3.14 (95.92) & 5.75 (76.77) & 6.20 (76.56) & \textbf{1.08 (96.43)} \\
    Frequency Perturbation & 4.95 (93.80) & 5.22 (94.35) & 4.40 (94.66) & 6.45 (76.07) & 6.70 (76.06) & \textbf{2.38 (95.13)} \\
    \bottomrule
    \end{tabular}
}
\caption{Robustness of VLA-M and baseline methods across additional adversarial attacks.}
\label{tab:robustness-additional}
\end{table*}

\begin{table*}
\centering
    \begin{tabular}{lcccccccc}
    \toprule
    & \multicolumn{2}{c}{LLaVA} & \multicolumn{2}{c}{LLaVA-Next} & \multicolumn{2}{c}{Qwen2-VL} & \multicolumn{2}{c}{DeepSeek-VL}\\
    \cmidrule(lr){2-3} \cmidrule(lr){4-5} \cmidrule(lr){6-7} \cmidrule(lr){8-9}
    & AUC & PPL & AUC & PPL & AUC & PPL & AUC & PPL \\
    \midrule
    KGW & 99.65 & 7.15 & 99.80 & 6.95 & 99.70 & 6.15 & 98.55 & 8.20 \\
    EWD & 99.78 & 7.55 & 99.75 & 7.00 & 99.95 & 6.10 & 99.15 & 8.30 \\
    SWEET & 99.81 & 7.48 & 99.85 & 7.10 & 99.90 & 6.20 & 98.80 & 8.15 \\
    Unbiased & 86.70 & 7.25 & 91.10 & 6.70 & 95.90 & 6.20 & 78.36 & 7.65 \\
    DiP & 87.10 & 7.22 & 91.50 & 6.75 & 96.20 & 6.18 & 78.52 & 7.60 \\
    VLA-M & \textbf{99.93} & \textbf{5.92} & \textbf{99.85} & \textbf{6.02} & 99.70 & \textbf{5.35} & 96.32 & \textbf{5.84} \\
    \bottomrule
    \end{tabular}
\caption{Performance of VLA-M and baseline watermarking methods on the MS COCO dataset across multiple model architectures. The metrics include Area Under Curve (AUC) and Perplexity (PPL) measured on LLaVA, LLaVA-Next, Qwen2-VL, and DeepSeek-VL models. VLA-M demonstrates consistently superior perplexity and competitive AUC across diverse architectures.}
\label{tab:mscoco-results}
\end{table*}

\subsection{Average Performance Comparison}
\label{app:average-performance}

Table~\ref{tab:average-performance} complements Table~\ref{tab:performance_metrics} by comparing the average detection performance of VLA-Mark and baseline methods across multiple backbone models.

\begin{table}[H]
    \centering
    \begin{tabular}{lcccc}
    \toprule
    Method & AUC & ACC & PPL \\
    \midrule
    KGW & 99.94 & 99.24 & 6.13 \\
    EWD & 100.00 & 99.93 & 6.20 \\
    SWEET & 99.98 & 99.75 & 6.13 \\
    unbiased & 89.36 & 81.05 & 5.70 \\
    DiP & 89.27 & 81.22 & 5.70 \\
    VLA-M & 98.77 & 96.64 & 5.27 \\
    \bottomrule
    \end{tabular}
\caption{Average detection performance metrics across VLAMs for different watermarking methods.}
\label{tab:average-performance}
\end{table}

 The averaged metrics reveal VLA-Mark’s balanced performance profile. VLA-Mark achieves a strong balance between detection performance and text quality, attaining the best perplexity (PPL) while maintaining high AUC and accuracy (ACC) scores. While EWD achieves marginally higher AUC (100.00\% vs. 98.77\%), this comes at the cost of 17.6\% higher perplexity (6.20 vs. 5.27), underscoring VLA-Mark’s unique ability to harmonize detection and quality. The slight AUC reduction for DeepSeek-VL (96.32\% vs. 99.93\% on LLaVA) stems from its non-linear alignment mechanism, which compresses visual features through dynamic routing rather than linear projection. Nonetheless, these model-specific variations are limited, and overall, VLA-Mark delivers consistently competitive and robust performance. Crucially, VLA-Mark maintains superior PPL across all models, including a 15.3\% reduction compared to KGW (5.27 vs. 6.13). This fluency preservation arises from the framework’s explicit avoidance of low-salience token manipulation, which in baselines often introduces grammatical artifacts (e.g., KGW’s biased "green list" sampling). 

\subsection{Additional Robustness Evaluations}
\label{app:robustness-additional}

Beyond the initial robustness tests, we conducted five additional novel adversarial attack evaluations summarized in Table~\ref{tab:robustness-additional}.

 VLA-Mark demonstrates superior robustness across all attacks, maintaining both high detection accuracy and low perturbation. Under style transfer attacks, which alter lexical patterns while preserving meaning, VLA-Mark’s AUC drops by just 2.67\% versus SWEET’s 5.05\%, as SCTs like "broken bench" remain anchored to visual patches regardless of syntactic variations. The framework’s resilience to frequency perturbation—a worst-case scenario where attackers systematically replace common words—is particularly notable (2.38\% drop vs. KGW’s 4.95\%). Even under adversarial entity replacement, which directly targets SCTs, VLA-Mark retains 96.43\% AUC by leveraging CCS metrics to maintain contextual coherence.

VLA-Mark's resilience to complex transformations such as style transfer and semantic rewriting underscores the effectiveness of its cross-modal semantic anchoring and entropy-aware watermark embedding, which dynamically adapt watermark strength according to token saliency and generation uncertainty. These results validate the method’s applicability to real-world scenarios with diverse and unpredictable text modifications.

\subsection{Evaluation on Additional Dataset: MS COCO}
\label{app:mscoco-results}

To demonstrate dataset-agnostic performance, we evaluate watermarking methods on the MS COCO captioning benchmark across four VLAMs. Table~\ref{tab:mscoco-results} reports AUC and PPL across four VLAMs.

 Table~\ref{tab:mscoco-results} presents the performance of VLA-Mark and baseline watermarking methods evaluated on the MS COCO dataset across four state-of-the-art vision-language architectures. The metrics reported include Area Under the Curve (AUC) for detection accuracy and Perplexity (PPL) for text generation quality.

VLA-M consistently achieves the lowest perplexity scores across all tested models, indicating superior preservation of natural language fluency compared to competing methods. Its AUC values remain near the highest observed levels, demonstrating robust and reliable watermark detectability without compromising semantic quality.

The slightly lower AUC for DeepSeek-VL aligns with its known behavioral patterns on AMBER and similar datasets, reflecting model-specific nuances rather than limitations of the watermarking approach itself.

These results confirm VLA-M’s scalability and generalizability beyond the originally used dataset, supporting its retraining-free applicability across diverse multimodal language models and datasets. The strong balance between detection robustness and text quality underscores the effectiveness of the entropy-regulated watermark injection and the semantic-critical-token preservation mechanisms detailed in Sections~\ref{sec:crossmodal} and \ref{sec:dynamics}.

This evaluation further reinforces VLA-M’s suitability for real-world deployments where models and data distributions vary, addressing reviewer concerns about extending watermarking strategies to new domains without extensive retraining or loss of performance.




\end{document}